\documentclass{article}
\usepackage[T1]{fontenc}
\usepackage{iclr2027_conference,times}
\usepackage{amsmath,amssymb}
\usepackage{array,booktabs,multirow}
\usepackage{graphicx}
\usepackage{float}
\floatstyle{ruled}
\newfloat{algorithm}{tbp}{loa}
\floatname{algorithm}{Algorithm}
\usepackage{flafter}
\usepackage{placeins}
\usepackage{needspace}
\usepackage{etoolbox}
\usepackage{microtype}
\usepackage{hyperref}
\usepackage{url}
\usepackage{xcolor}
\usepackage{colortbl}
\definecolor{carryshade}{RGB}{233,239,246}
\usepackage{tabularx}
\hypersetup{hidelinks}

\setkeys{Gin}{keepaspectratio}
\title{Carryover Drafting: Recycling Rejected \\ States for Speculative Decoding}

\author{
Jahyun Koo$^{1\text{*}}$ \qquad \hspace{0.32cm}Sunghyeon Woo$^{2}$ \qquad \hspace{0.32cm} Jaeeun Kil$^{2}$ \hspace{0.32cm} \qquad Jeongtae Lee$^{2}$ \\
{\hspace{-0.05em} \bf Sungjae Lee$^{2}$ \hspace{1.01cm} Kyomin Jung$^{1\dagger}$ \qquad \ \hspace{0.59cm}Minsub Kim$^{2\dagger}$} \\[0.5em]
\hspace{-0.05em} $^{1}$Seoul National University \qquad $^{2}$NAVER Cloud
}

\iclrfinalcopy

\begin{document}
\raggedbottom
\maketitle

\begingroup
\renewcommand{\thefootnote}{\fnsymbol{footnote}}
\footnotetext[1]{Work done during an internship at NAVER Cloud.}
\footnotetext[2]{Corresponding authors.}
\endgroup
\lhead{Under review as a conference paper at ICLR 2027}

\begin{abstract}
Speculative decoding accelerates LLM inference by verifying multiple drafted tokens in parallel, allowing a single target forward pass to accept several tokens. By construction, verification computes representations for both accepted and rejected tokens. Yet, conventional drafters retain only the representations of accepted tokens, leaving the substantial verifier computation spent on rejected tokens effectively wasted. We find that these discarded hidden states generated during target forward retain useful information about future tokens that can improve subsequent drafts. However, realizing this opportunity poses two distinct challenges. At inference, recycling overhead can increase drafting latency, diminishing the speedup gained from increased acceptance length. During training, standard parallel drafter training does not produce inference-aligned rejected states, while obtaining them through sequential rollouts would sacrifice parallelism across training positions. We introduce \textbf{Carryover Drafting}, which addresses both challenges. Carryover recycles rejected target hidden states as temporary KV context, allowing the drafter to selectively attend to them. It reuses the drafter's existing interface and adds only a single learned embedding to distinguish rejected states from committed context. The additional KV context is replaced each drafting round, keeping its length bounded by one proposal block. We introduce parallel \emph{draft--verify--draft} training that exposes the drafter to inference-aligned rejected states while preserving parallelism across training positions. Experiments with DFlash and a DSpark-derived semi-autoregressive drafter across two target models show that this simple Carryover mechanism improves average acceptance length by 6.5--14.7\% and end-to-end vLLM speedup by 7.9--14.4\% over the corresponding baselines, with speedup gains reaching 28.8\% on translation.
\end{abstract}
\begin{figure}[H]
\centering
\includegraphics[width=0.92\textwidth]{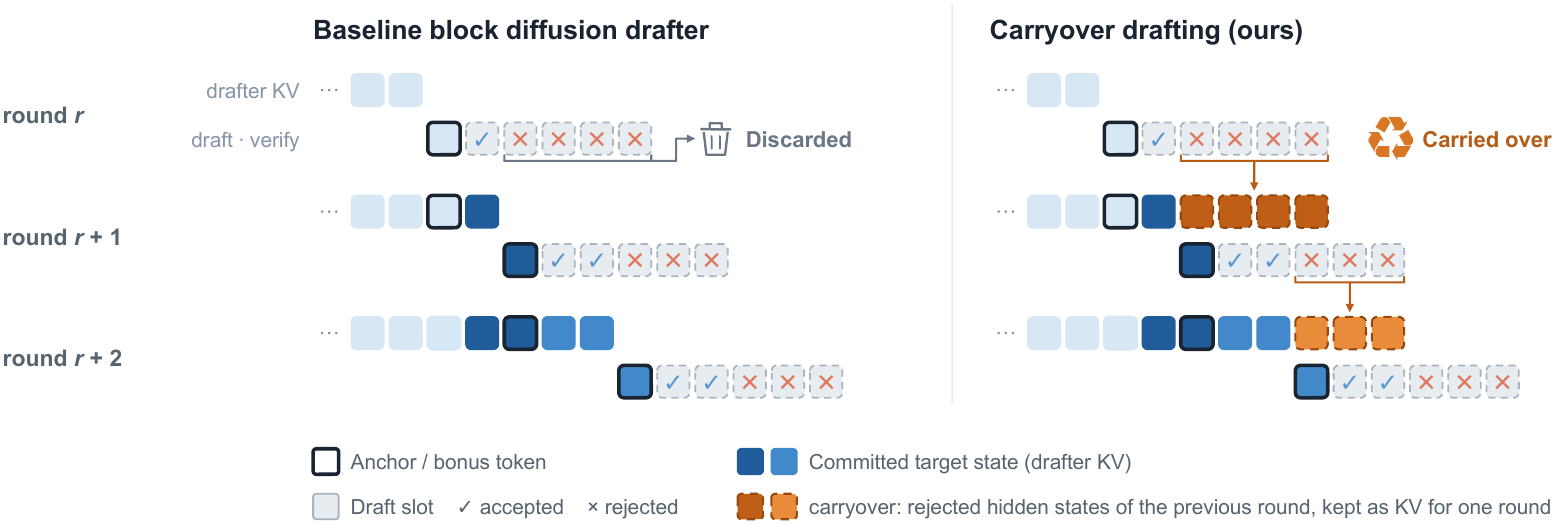}
\caption{\textbf{Carryover Drafting.} Standard block diffusion drafting retains accepted target hidden states and discards rejected states. Carryover recycles rejected states as temporary drafter KV context for the next proposal, allowing the drafter to selectively attend to them through its existing attention layers.}
\label{fig:overview}
\end{figure}

\newpage

\section{Introduction}

Autoregressive LLMs generate one token per forward pass. Speculative decoding uses additional parallel computation to accelerate this process: a lightweight drafter proposes several tokens, which the target verifies in a single forward pass, allowing multiple tokens to be accepted at once \citep{leviathan2023fast,chen2023accelerating}. Better drafts allow more tokens to be accepted per target pass, making draft quality a central factor in acceleration. To improve draft quality, recent methods reuse hidden states already computed by the target. EAGLE uses target hidden states immediately before the language modeling head, while EAGLE-3 combines states from multiple target layers \citep{li2024eagle,li2025eagle3}. Block diffusion drafters such as DFlash and DSpark similarly use hidden states from multiple target layers, projecting them into the drafter's KV context to improve the quality of parallel block prediction \citep{chen2026dflash,cheng2026dspark}.

However, these drafters retain hidden states from accepted tokens and discard those from rejected tokens. In speculative decoding, a draft is often only partially accepted. By construction, verification computes representations for both accepted and rejected tokens. This leaves substantial verifier computation spent on rejected tokens effectively wasted, raising a natural question: can these rejected hidden states be carried over to improve subsequent drafts?

Nevertheless, realizing this opportunity poses two distinct challenges. At inference, speedup from speculative decoding depends jointly on acceptance length and drafting overhead. The cost of projecting and attending to additional hidden states can outweigh the benefit of accepting more tokens, while retaining rejected states across rounds would lead to an ever-growing KV context. During training, the drafter must learn to use the rejected hidden states that arise at inference time when the target verifies the drafter's own proposals. However, standard parallel drafter training on recorded target responses does not expose these states. A straightforward way to obtain them is through sequential rollouts, but this would make later training positions depend on earlier draft-and-verify outcomes, sacrificing parallelism across positions.

We introduce \textbf{Carryover Drafting}, which addresses both challenges. Carryover recycles rejected target hidden states as temporary KV context for the next draft through the same projections used for accepted states (Figure~\ref{fig:overview}). This lets the drafter learn to selectively attend to carried-over states through its existing attention layers. Without adding a dedicated gate or projection module, Carryover requires only a single new learned embedding to distinguish rejected states from accepted context, keeping the recycling interface lightweight. We replace the carried-over states each drafting round, bounding the additional KV context by one proposal block regardless of how much context has accumulated. Carryover therefore requires only the addition and replacement of carried-over states within the existing drafting interface, simplifying integration with standard block diffusion drafters.

\begin{figure*}[t]
\centering
\includegraphics[width=0.90\textwidth]{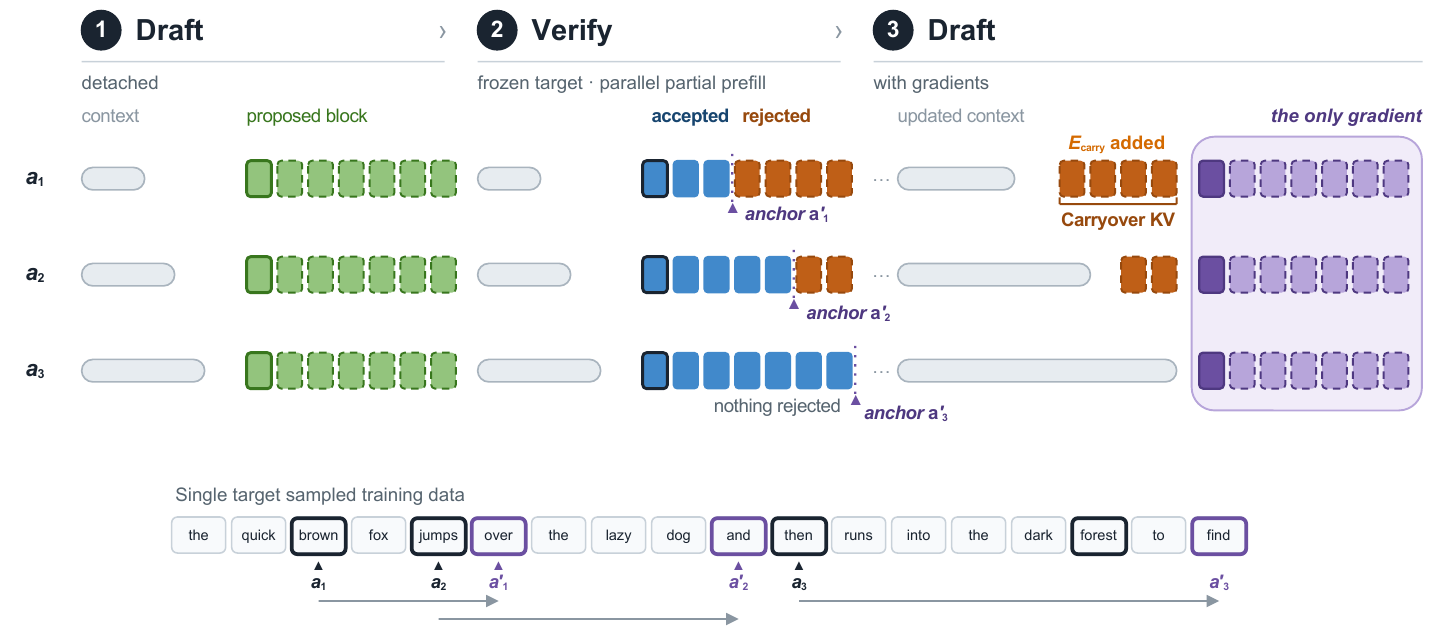}
\caption{\textbf{Parallel draft--verify--draft training.} Randomly sampled anchors share a single recorded target response. Detached drafter proposals are verified by a frozen target; the second draft uses rejected states and recorded supervision at shifted anchors.}
\label{fig:training}
\end{figure*}

To obtain inference-aligned rejected states without sequential rollouts, we introduce parallel \textit{draft--verify--draft} training (Figure~\ref{fig:training}). At sampled positions within a recorded target response, the drafter first generates detached proposals, which a frozen target verifies in parallel. Verification determines the next drafting position and produces the rejected hidden states that would naturally be available there at inference time. These states are then carried over to a subsequent draft, which is trained against the original recorded continuation at the shifted position, with gradients flowing only through this final drafting stage. Because each draft--verify--draft transition is constructed independently, training remains parallel across positions. This procedure does not augment the target-sampled training data or generate new target continuations: the recorded responses remain the source of supervision, and the intermediate draft--verify stage supplies the shifted positions and rejected hidden states needed for carryover training.

Across two target models and eight datasets, Carryover improves both DFlash and a DSpark-derived semi-autoregressive drafter, increasing average acceptance length by 6.5--14.7\% and end-to-end vLLM speedup by 7.9--14.4\% over the corresponding baselines. Surprisingly, we find speedup gains reaching up to 28.8\% on translation. These results demonstrate that recycling rejected hidden states already computed during inference can substantially improve subsequent drafts through a simple carryover mechanism.

\section{Carryover Drafting}

Carryover adds rejected target hidden states to the KV context for the next draft. We define the inference update and then show how to construct its training inputs in parallel.

\paragraph{Preliminaries.}
In each round, the drafter proposes $B$ tokens and the target accepts a prefix of length $m$. The target additionally supplies a token after the accepted prefix, which we refer to as the bonus token. Acceptance length $\tau$ includes that token. DFlash predicts a block from an anchor followed by masked positions \citep{chen2026dflash}. Hidden states from selected target layers are concatenated, fused by a projection $F$, and passed to the key and value projections of every drafter layer. Draft positions attend to keys and values computed from these target hidden states by the learned drafter projections. Our DFlash and Markov baselines share this interface, which Carryover reuses for rejected states.

\paragraph{Acceptance and the cost of reuse.}
Following the per-token latency model in DFlash and DSpark \citep{chen2026dflash,cheng2026dspark}, we account for the incremental cost of carryover. Latency $L_{\mathrm{carry}}$ and speedup $S_{\mathrm{carry}}$ are approximately
\[
L_{\mathrm{carry}}\approx\frac{t_{\mathrm{draft}}+t_{\mathrm{verify}}+t_{\mathrm{reuse}}}{\tau},\qquad
S_{\mathrm{carry}}\approx\frac{\tau\,t_{\mathrm{AR}}}{t_{\mathrm{draft}}+t_{\mathrm{verify}}+t_{\mathrm{reuse}}}.
\]
Here $t_{\mathrm{AR}}$ is autoregressive latency per token. With $t_{\mathrm{draft}}=T_{\mathrm{draft}}^{\mathrm{baseline}}$, we define $t_{\mathrm{reuse}}=T_{\mathrm{draft}}^{\mathrm{Carryover}}-T_{\mathrm{draft}}^{\mathrm{baseline}}$ as the additional drafting latency caused by recycling rejected hidden states. Speedup therefore requires higher acceptance with low overhead. This motivates a lightweight carryover design: reusing existing projections and replacing rejected context every round keeps the added latency of recycling rejected hidden states minimal.

\begin{samepage}
\subsection{One-round reuse through the existing interface}
Let $H_r=(h_{r,1},\ldots,h_{r,B})$ be the target hidden states at the proposed token positions in round $r$. Each $h_{r,j}$ concatenates the selected target-layer hidden states in the same order used by the baseline. If $m_r$ proposals are accepted, define
\begin{equation}
H_r^{\mathrm{acc}}=h_{r,1:m_r},\quad H_r^{\mathrm{rej}}=h_{r,m_r+1:B},\quad
\Phi_\ell(H)=\bigl(K_\ell(F(H)),V_\ell(F(H))\bigr),
\label{eq:features}
\end{equation}
where $F$ is the existing projection and $K_\ell,V_\ell$ are the existing projections in drafter layer $\ell$. The equations show feature and KV projections; attention normalization and positional encoding are omitted. Let $C_{r+1,\ell}$ denote committed drafter KV context after the usual baseline update. Carryover constructs
\begin{equation}
U_{r+1,\ell}=\Phi_\ell(H_r^{\mathrm{rej}}+E_{\mathrm{carry}}),\qquad
\mathrm{ctx}_{r+1,\ell}=[C_{r+1,\ell};U_{r+1,\ell}].
\label{eq:carryover}
\end{equation}
\end{samepage}
\paragraph{A lightweight recycling interface.}
The embedding $E_{\mathrm{carry}}$ distinguishes rejected states from accepted context before projection by $F$. This embedding is the only module added to the baseline drafter; the drafter and its existing projections are also fine-tuned. A dedicated projector gives slightly higher acceptance in our interface probe (Section~\ref{sec:interface}), but introduces another projection module on the drafting path.

Rejected states serve as additional read-only drafter context. They neither replace the hidden states of accepted tokens nor enter the target KV cache. Each verification replaces the previous carried-over context, and full acceptance leaves it empty. The added context is bounded by $B$ rows per drafter layer, with storage reused from the verification window. For the full inference procedure and cache semantics, see Algorithm~\ref{alg:inference} in Appendix~\ref{app:algorithms}.

\subsection{Parallel training on recorded target responses}
Following parallel block-drafter training in DFlash and DSpark, we sample random anchor positions $a_i$ along each recorded target response $y$ \citep{chen2026dflash,cheng2026dspark}. Baseline and Carryover use identical target-sampled training data, with 512 anchors per response, and the same baseline-specific loss. This includes exponential position weighting, which prioritizes earlier draft positions, since later tokens can be accepted only if all preceding draft tokens are accepted. Appendix~\ref{app:training-settings} gives the loss components.

\Needspace{7\baselineskip}
\paragraph{Parallel draft--verify--draft.}
For Carryover, we construct a local transition at each sampled anchor. The first drafter pass generates proposals without gradients. A frozen target verifies the branches in parallel using a block-diagonal causal mask: each branch reads its recorded prefix and its own proposed predecessors. Greedy verification returns accepted lengths $m_i$ and detached rejected hidden states $H_i^{\mathrm{rej}}$.

Each anchor shifts to $a'_i=a_i+m_i+1$, the next correction or bonus position. The second drafter pass reads the recorded prefix through $a'_i$, receives $H_i^{\mathrm{rej}}$ through Equation~\ref{eq:carryover}, and learns the recorded continuation. When all tokens are accepted, no rejected states are carried over, and the second drafter pass becomes equivalent to baseline training at the shifted anchor. We retain such cases in training.

Gradients flow only through the second drafter pass, including the shared projections and carryover embedding; the first pass and target verification are detached. Each sampled anchor supplies 1 supervised transition on the same recorded response. Training remains parallel across anchors without a full multi-round speculative rollout for every training example. For the full training procedure, see Algorithm~\ref{alg:training} in Appendix~\ref{app:algorithms}.

\begin{table*}[t]
\definecolor{carryshade}{RGB}{233,239,246}
\centering
\caption{\textbf{Acceptance and speedup over autoregressive decoding.} Per-dataset speedup uses offline-vLLM, and Avg. speedup is a separate online-vLLM  measurement. Avg. acceptance is micro average over 844 prompts.}
\label{tab:big}
\scriptsize
\setlength{\tabcolsep}{4pt}
\resizebox{\textwidth}{!}{%
\begin{tabular}{lll r@{\,$|$\,}>{\columncolor{carryshade}[0pt][\tabcolsep]}l r@{\,$|$\,}>{\columncolor{carryshade}[0pt][\tabcolsep]}l r@{\,$|$\,}>{\columncolor{carryshade}[0pt][\tabcolsep]}l r@{\,$|$\,}>{\columncolor{carryshade}[0pt][\tabcolsep]}l}
\toprule
 & & & \multicolumn{4}{c}{$T{=}0$} & \multicolumn{4}{c}{$T{=}1$} \\
\cmidrule(lr){4-7}\cmidrule(lr){8-11}
 & & & \multicolumn{2}{c}{$\tau$} & \multicolumn{2}{c}{Speedup} & \multicolumn{2}{c}{$\tau$} & \multicolumn{2}{c}{Speedup} \\
\cmidrule(lr){4-5}\cmidrule(lr){6-7}\cmidrule(lr){8-9}\cmidrule(lr){10-11}
Target & Drafter & Dataset & Baseline & \textbf{+Carryover} & Baseline & \textbf{+Carryover} & Baseline & \textbf{+Carryover} & Baseline & \textbf{+Carryover} \\
\midrule
 \multirow{18}{*}{Qwen3-4B} & \multirow{9}{*}{DFlash} & HumanEval & 4.89 & \textbf{5.33}{\tiny(+9.0\%)} & 4.74$\times$ & \textbf{5.17$\times$}{\tiny(+9.0\%)} & 4.22 & \textbf{4.53}{\tiny(+7.4\%)} & 3.97$\times$ & \textbf{4.53$\times$}{\tiny(+14.0\%)} \\
  &  & Math & 5.96 & \textbf{6.51}{\tiny(+9.2\%)} & 5.58$\times$ & \textbf{6.10$\times$}{\tiny(+9.4\%)} & 5.31 & \textbf{5.64}{\tiny(+6.1\%)} & 5.09$\times$ & \textbf{5.34$\times$}{\tiny(+4.9\%)} \\
  &  & MT-Bench & 3.45 & \textbf{3.77}{\tiny(+9.1\%)} & 3.40$\times$ & \textbf{3.72$\times$}{\tiny(+9.5\%)} & 2.98 & \textbf{3.18}{\tiny(+6.5\%)} & 2.72$\times$ & \textbf{2.95$\times$}{\tiny(+8.6\%)} \\
  &  & QA & 3.03 & \textbf{3.26}{\tiny(+7.7\%)} & 2.96$\times$ & \textbf{3.15$\times$}{\tiny(+6.5\%)} & 2.61 & \textbf{2.84}{\tiny(+8.9\%)} & 2.60$\times$ & \textbf{2.64$\times$}{\tiny(+1.5\%)} \\
  &  & RAG & 3.19 & \textbf{3.48}{\tiny(+9.2\%)} & 3.19$\times$ & \textbf{3.49$\times$}{\tiny(+9.5\%)} & 2.96 & \textbf{2.99}{\tiny(+1.1\%)} & 2.75$\times$ & \textbf{2.83$\times$}{\tiny(+3.1\%)} \\
  &  & Summ. & 2.83 & \textbf{3.15}{\tiny(+11.3\%)} & 2.72$\times$ & \textbf{2.97$\times$}{\tiny(+9.3\%)} & 2.52 & \textbf{2.72}{\tiny(+8.1\%)} & 2.63$\times$ & \textbf{2.80$\times$}{\tiny(+6.6\%)} \\
  &  & Tool & 3.50 & \textbf{3.84}{\tiny(+10.0\%)} & 3.39$\times$ & \textbf{3.78$\times$}{\tiny(+11.5\%)} & 3.13 & \textbf{3.40}{\tiny(+8.6\%)} & 3.11$\times$ & \textbf{3.20$\times$}{\tiny(+2.8\%)} \\
  &  & Transl. & 2.57 & \textbf{3.14}{\tiny(+22.1\%)} & 2.12$\times$ & \textbf{2.41$\times$}{\tiny(+13.8\%)} & 2.26 & \textbf{2.67}{\tiny(+18.0\%)} & 1.89$\times$ & \textbf{2.18$\times$}{\tiny(+15.3\%)} \\
\cmidrule(lr){3-11}
  &  & \textbf{Avg.} & 3.78 & \textbf{4.14}{\tiny(+9.4\%)} & 3.34$\times$ & \textbf{3.60$\times$}{\tiny(+7.9\%)} & 3.33 & \textbf{3.55}{\tiny(+6.5\%)} & 2.89$\times$ & \textbf{3.14$\times$}{\tiny(+8.6\%)} \\
\cmidrule(lr){2-11}
  & \multirow{9}{*}{Markov} & HumanEval & 5.75 & \textbf{6.30}{\tiny(+9.6\%)} & 5.02$\times$ & \textbf{5.47$\times$}{\tiny(+9.0\%)} & 5.36 & \textbf{5.69}{\tiny(+6.2\%)} & 4.72$\times$ & \textbf{4.87$\times$}{\tiny(+3.1\%)} \\
  &  & Math & 7.29 & \textbf{7.91}{\tiny(+8.4\%)} & 6.03$\times$ & \textbf{6.52$\times$}{\tiny(+8.1\%)} & 6.63 & \textbf{7.42}{\tiny(+11.8\%)} & 5.54$\times$ & \textbf{6.15$\times$}{\tiny(+11.0\%)} \\
  &  & MT-Bench & 3.91 & \textbf{4.37}{\tiny(+11.6\%)} & 3.47$\times$ & \textbf{3.84$\times$}{\tiny(+10.7\%)} & 3.68 & \textbf{4.08}{\tiny(+10.8\%)} & 3.09$\times$ & \textbf{3.67$\times$}{\tiny(+18.9\%)} \\
  &  & QA & 3.42 & \textbf{3.80}{\tiny(+11.2\%)} & 2.98$\times$ & \textbf{3.30$\times$}{\tiny(+10.8\%)} & 3.32 & \textbf{3.45}{\tiny(+4.0\%)} & 2.75$\times$ & \textbf{2.98$\times$}{\tiny(+8.5\%)} \\
  &  & RAG & 3.64 & \textbf{4.07}{\tiny(+11.9\%)} & 3.12$\times$ & \textbf{3.47$\times$}{\tiny(+11.1\%)} & 3.44 & \textbf{3.79}{\tiny(+10.1\%)} & 3.00$\times$ & \textbf{3.30$\times$}{\tiny(+9.9\%)} \\
  &  & Summ. & 3.14 & \textbf{3.63}{\tiny(+15.4\%)} & 2.66$\times$ & \textbf{3.06$\times$}{\tiny(+15.0\%)} & 2.99 & \textbf{3.41}{\tiny(+13.9\%)} & 2.68$\times$ & \textbf{3.08$\times$}{\tiny(+14.8\%)} \\
  &  & Tool & 3.96 & \textbf{4.49}{\tiny(+13.5\%)} & 3.48$\times$ & \textbf{3.93$\times$}{\tiny(+12.7\%)} & 3.70 & \textbf{4.22}{\tiny(+13.8\%)} & 3.23$\times$ & \textbf{3.56$\times$}{\tiny(+10.5\%)} \\
  &  & Transl. & 2.84 & \textbf{3.61}{\tiny(+26.8\%)} & 2.00$\times$ & \textbf{2.54$\times$}{\tiny(+27.3\%)} & 2.65 & \textbf{3.41}{\tiny(+28.6\%)} & 1.94$\times$ & \textbf{2.50$\times$}{\tiny(+28.8\%)} \\
\cmidrule(lr){3-11}
  &  & \textbf{Avg.} & 4.33 & \textbf{4.85}{\tiny(+11.8\%)} & 3.44$\times$ & \textbf{3.83$\times$}{\tiny(+11.5\%)} & 4.06 & \textbf{4.48}{\tiny(+10.3\%)} & 3.20$\times$ & \textbf{3.54$\times$}{\tiny(+10.6\%)} \\
\midrule
 \multirow{18}{*}{Gemma 4 12B} & \multirow{9}{*}{DFlash} & HumanEval & 4.86 & \textbf{5.28}{\tiny(+8.7\%)} & 3.65$\times$ & \textbf{4.02$\times$}{\tiny(+10.0\%)} & 4.23 & \textbf{4.50}{\tiny(+6.3\%)} & 2.98$\times$ & \textbf{3.18$\times$}{\tiny(+6.7\%)} \\
  &  & Math & 5.44 & \textbf{6.03}{\tiny(+10.8\%)} & 3.97$\times$ & \textbf{4.35$\times$}{\tiny(+9.6\%)} & 4.70 & \textbf{5.24}{\tiny(+11.5\%)} & 3.49$\times$ & \textbf{3.94$\times$}{\tiny(+13.1\%)} \\
  &  & MT-Bench & 3.25 & \textbf{3.57}{\tiny(+10.1\%)} & 2.49$\times$ & \textbf{2.66$\times$}{\tiny(+7.0\%)} & 2.85 & \textbf{3.09}{\tiny(+8.2\%)} & 2.18$\times$ & \textbf{2.38$\times$}{\tiny(+9.5\%)} \\
  &  & QA & 2.87 & \textbf{3.15}{\tiny(+9.8\%)} & 2.23$\times$ & \textbf{2.44$\times$}{\tiny(+9.1\%)} & 2.41 & \textbf{2.57}{\tiny(+6.9\%)} & 1.84$\times$ & \textbf{1.96$\times$}{\tiny(+6.8\%)} \\
  &  & RAG & 3.63 & \textbf{3.99}{\tiny(+9.9\%)} & 2.68$\times$ & \textbf{2.93$\times$}{\tiny(+9.7\%)} & 3.27 & \textbf{3.62}{\tiny(+10.8\%)} & 2.41$\times$ & \textbf{2.73$\times$}{\tiny(+13.1\%)} \\
  &  & Summ. & 2.72 & \textbf{3.04}{\tiny(+12.0\%)} & 2.14$\times$ & \textbf{2.34$\times$}{\tiny(+9.4\%)} & 2.47 & \textbf{2.70}{\tiny(+9.3\%)} & 1.96$\times$ & \textbf{2.07$\times$}{\tiny(+5.6\%)} \\
  &  & Tool & 2.94 & \textbf{3.24}{\tiny(+10.3\%)} & 2.16$\times$ & \textbf{2.37$\times$}{\tiny(+9.9\%)} & 2.78 & \textbf{3.06}{\tiny(+10.1\%)} & 2.16$\times$ & \textbf{2.30$\times$}{\tiny(+6.7\%)} \\
  &  & Transl. & 3.14 & \textbf{3.60}{\tiny(+14.6\%)} & 2.44$\times$ & \textbf{2.75$\times$}{\tiny(+12.5\%)} & 2.78 & \textbf{3.15}{\tiny(+13.0\%)} & 2.11$\times$ & \textbf{2.44$\times$}{\tiny(+15.7\%)} \\
\cmidrule(lr){3-11}
  &  & \textbf{Avg.} & 3.65 & \textbf{4.03}{\tiny(+10.3\%)} & 2.58$\times$ & \textbf{2.79$\times$}{\tiny(+8.1\%)} & 3.24 & \textbf{3.52}{\tiny(+8.6\%)} & 2.40$\times$ & \textbf{2.59$\times$}{\tiny(+7.9\%)} \\
\cmidrule(lr){2-11}
  & \multirow{9}{*}{Markov} & HumanEval & 5.82 & \textbf{6.43}{\tiny(+10.3\%)} & 4.18$\times$ & \textbf{4.57$\times$}{\tiny(+9.2\%)} & 5.46 & \textbf{5.98}{\tiny(+9.5\%)} & 3.74$\times$ & \textbf{4.03$\times$}{\tiny(+7.8\%)} \\
  &  & Math & 6.41 & \textbf{7.18}{\tiny(+12.1\%)} & 4.37$\times$ & \textbf{4.85$\times$}{\tiny(+10.9\%)} & 6.06 & \textbf{6.87}{\tiny(+13.2\%)} & 4.34$\times$ & \textbf{4.64$\times$}{\tiny(+6.9\%)} \\
  &  & MT-Bench & 3.59 & \textbf{4.17}{\tiny(+16.2\%)} & 2.60$\times$ & \textbf{2.99$\times$}{\tiny(+15.2\%)} & 3.44 & \textbf{3.92}{\tiny(+14.0\%)} & 2.46$\times$ & \textbf{2.87$\times$}{\tiny(+16.7\%)} \\
  &  & QA & 3.14 & \textbf{3.64}{\tiny(+15.9\%)} & 2.31$\times$ & \textbf{2.65$\times$}{\tiny(+15.0\%)} & 2.81 & \textbf{3.40}{\tiny(+21.0\%)} & 2.06$\times$ & \textbf{2.22$\times$}{\tiny(+7.6\%)} \\
  &  & RAG & 4.16 & \textbf{4.73}{\tiny(+13.7\%)} & 2.89$\times$ & \textbf{3.22$\times$}{\tiny(+11.5\%)} & 3.96 & \textbf{4.35}{\tiny(+10.0\%)} & 2.66$\times$ & \textbf{2.94$\times$}{\tiny(+10.7\%)} \\
  &  & Summ. & 3.01 & \textbf{3.51}{\tiny(+16.6\%)} & 2.22$\times$ & \textbf{2.54$\times$}{\tiny(+14.4\%)} & 2.91 & \textbf{3.34}{\tiny(+15.0\%)} & 2.16$\times$ & \textbf{2.52$\times$}{\tiny(+16.8\%)} \\
  &  & Tool & 3.25 & \textbf{3.75}{\tiny(+15.2\%)} & 2.26$\times$ & \textbf{2.54$\times$}{\tiny(+12.5\%)} & 3.29 & \textbf{3.69}{\tiny(+12.2\%)} & 2.34$\times$ & \textbf{2.63$\times$}{\tiny(+12.4\%)} \\
  &  & Transl. & 3.51 & \textbf{4.18}{\tiny(+19.1\%)} & 2.55$\times$ & \textbf{3.02$\times$}{\tiny(+18.2\%)} & 3.22 & \textbf{3.96}{\tiny(+23.2\%)} & 2.30$\times$ & \textbf{2.78$\times$}{\tiny(+21.1\%)} \\
\cmidrule(lr){3-11}
  &  & \textbf{Avg.} & 4.13 & \textbf{4.74}{\tiny(+14.7\%)} & 2.73$\times$ & \textbf{3.13$\times$}{\tiny(+14.4\%)} & 3.92 & \textbf{4.49}{\tiny(+14.5\%)} & 2.71$\times$ & \textbf{3.09$\times$}{\tiny(+14.3\%)} \\
\bottomrule
\end{tabular}
}
\end{table*}

\section{Experimental setup}
\label{sec:setup}
\paragraph{Targets and drafters.}
We evaluate Qwen3-4B and Gemma 4 12B \citep{yang2025qwen3,gemmateam2026gemma4} with 2 drafting regimes. \textbf{DFlash} predicts a block in parallel using target hidden states \citep{chen2026dflash}. \textbf{Markov} adds the rank-256 first-order correction head introduced in DSpark to the parallel backbone \citep{cheng2026dspark}. It represents semi-autoregressive block drafting within the broader effort to strengthen within-block modeling, including Domino, xPress, and DFlash 2 \citep{huang2026domino,wang2026xpress,incoai2026dflash2}. Comparing these regimes tests whether carryover remains beneficial with a stronger drafter that models dependencies between proposed tokens. Both use 5 drafter layers and 15 proposed tokens; confidence scheduling is disabled to keep verification length fixed.

\paragraph{Training configuration.}
All variants use the same 500K prompts from UltraChat-200K \citep{ding2023ultrachat} and Magpie-Llama-3.1-Pro-300K-Filtered \citep{xu2024magpie}, following the public speculator data recipe documented by \citet{amazon2026peaglecard}. Each target regenerates the responses, which its Baseline and Carryover variants share. We train all baselines for 10 epochs following DSpark. This choice follows the convergence rationale reported in DSpark: all compared drafters, including DFlash, were trained for 10 epochs to ensure full convergence \citep{cheng2026dspark}. This exceeds the 6 epochs reported in the original DFlash recipe \citep{chen2026dflash}. Carryover then trains for 3 epochs from the 10-epoch baseline checkpoint, using the same recorded responses. Appendix~\ref{app:training-settings} gives the shared training settings.

\paragraph{Evaluation.}
The Red Hat speculator benchmark contains 844 prompts: 164 for HumanEval, 200 for tool calling, and 80 each for Math, MT-Bench, QA, RAG, summarization, and translation \citep{redhat2026speculators}. We evaluate at $T=0$ and $T=1$ using vLLM on 1 H200 GPU \citep{kwon2023vllm}. Acceptance length includes bonus tokens from verification. Baseline--Carryover pairs share evaluation settings. Average speedup is measured on the combined benchmark workload, separately from dataset speedups; Appendix~\ref{app:evaluation} specifies workloads and aggregation.

\section{Results}
\begin{figure}[t]
\centering
\includegraphics[width=0.96\textwidth]{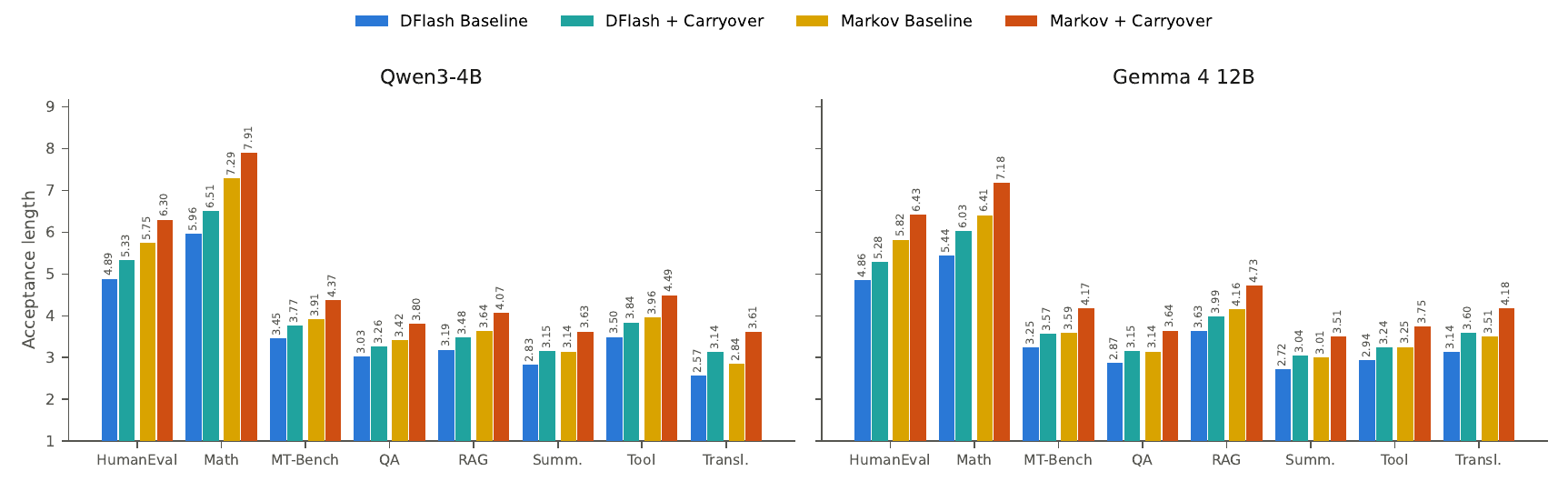}
\caption{\textbf{Acceptance length at $T=0$.} Baseline and Carryover for DFlash and Markov with Qwen3-4B and Gemma 4 12B across 8 datasets. The full measurements are in Table~\ref{tab:big}.}
\label{fig:acceptance}
\end{figure}
\subsection{Acceptance across tasks and drafting regimes}
We first test whether rejected hidden states improve both drafting regimes across targets. Table~\ref{tab:big} shows acceptance gains in every evaluated dataset setting, with average acceptance-length improvements of 6.5--14.7\%. The gains persist when moving from fully parallel DFlash to semi-autoregressive Markov. Carryover therefore complements the within-block dependency modeling of a stronger drafter.

The task breakdown reveals where recycling helps most. Translation has the largest relative acceptance gain in every target--drafter--temperature combination, reaching 28.6\% for Qwen3-4B Markov at $T=1$. This variation motivates the rejected-branch analysis in Section~\ref{sec:reconvergence}: does carryover help more when discarded states retain more recoverable information? Figure~\ref{fig:acceptance} compares acceptance length across datasets at $T=0$; Appendix~\ref{app:acceptance-t1} reports $T=1$.

\subsection{End-to-end speedup and throughput}
Higher acceptance is useful only if it compensates for the cost of recycling. The average speedup measurements in Table~\ref{tab:big} include projection and attention over carried-over states: Carryover improves speedup over autoregressive decoding by 7.9--14.4\% relative to the corresponding baseline. Thus the acceptance gains survive the added drafting overhead.

To evaluate speedup across serving loads, we vary concurrency from 1 to 32 (Table~\ref{tab:conc}). As concurrency increases, both Baseline and Carryover provide smaller speedups over autoregressive decoding, but Carryover maintains relative gains over Baseline for both targets and drafting regimes. These experiments use fixed verification lengths. DSpark instead uses confidence scheduling to reduce verification cost and sustain speedup at higher concurrency \citep{cheng2026dspark}. Section~\ref{sec:future} discusses how carryover could complement such adaptive scheduling.
\begin{table*}[t]
\definecolor{carryshade}{RGB}{233,239,246}
\providecommand{\cogain}[2]{\makebox[3.05em][r]{\textbf{#1}}{\tiny(+#2\%)}}
\centering
\caption{\textbf{Throughput across concurrency.} Output tokens/s on the mixed 844-prompt workload at \(T=0\) on a single H200 GPU. }
\label{tab:conc}
\small
\setlength{\tabcolsep}{5pt}
\begin{tabular}{c r r r@{\,$|$\,}>{\columncolor{carryshade}[0pt][\tabcolsep]}l r@{\,$|$\,}>{\columncolor{carryshade}[0pt][\tabcolsep]}l}
\toprule
 & & \multicolumn{5}{c}{Throughput (tok/s)} \\
\cmidrule(lr){3-7}
Target & $c$ & AR & \multicolumn{2}{c}{DFlash} & \multicolumn{2}{c}{Markov} \\
\cmidrule(lr){4-5}\cmidrule(lr){6-7}
 & & & Baseline & \textbf{+Carryover} & Baseline & \textbf{+Carryover} \\
\midrule
\multirow{6}{*}{Qwen3-4B}
 & 1  & 251   & 837   & \cogain{903}{7.9}   & 861   & \cogain{961}{11.5} \\
 & 2  & 491   & 1,609 & \cogain{1,739}{8.1} & 1,652 & \cogain{1,840}{11.4} \\
 & 4  & 946   & 3,015 & \cogain{3,275}{8.6} & 3,097 & \cogain{3,443}{11.2} \\
 & 8  & 1,800 & 5,328 & \cogain{5,741}{7.7} & 5,337 & \cogain{5,879}{10.1} \\
 & 16 & 3,181 & 7,243 & \cogain{7,785}{7.5} & 7,499 & \cogain{8,269}{10.3} \\
 & 32 & 5,335 & 8,180 & \cogain{8,804}{7.6} & 8,608 & \cogain{9,402}{9.2} \\
\midrule
\multirow{6}{*}{Gemma 4 12B}
 & 1  & 111   & 286   & \cogain{309}{8.2}   & 304   & \cogain{347}{14.4} \\
 & 2  & 216   & 553   & \cogain{594}{7.3}   & 584   & \cogain{671}{14.9} \\
 & 4  & 419   & 1,035 & \cogain{1,109}{7.2} & 1,097 & \cogain{1,238}{12.9} \\
 & 8  & 785   & 1,821 & \cogain{1,935}{6.3} & 1,908 & \cogain{2,145}{12.4} \\
 & 16 & 1,388 & 2,455 & \cogain{2,650}{8.0} & 2,573 & \cogain{2,892}{12.4} \\
 & 32 & 2,297 & 2,789 & \cogain{3,000}{7.6} & 2,973 & \cogain{3,315}{11.5} \\
\bottomrule
\end{tabular}
\end{table*}

\subsection{Carryover interface ablation}
\label{sec:interface}

\begin{table}[t]
\definecolor{carryshade}{RGB}{233,239,246}

\providecommand{\abcellH}[2]{%
  \makebox[3.5em][r]{#1}%
  \makebox[2.5em][l]{\,{\tiny(#2\%)}}%
}
\providecommand{\abcellboldH}[2]{%
  \makebox[3.5em][r]{\textbf{#1}}%
  \makebox[2.5em][l]{\,{\tiny(#2\%)}}%
}
\providecommand{\abbaseH}[1]{%
  \makebox[3.5em][r]{#1}%
  \makebox[2.5em][l]{\,{}}%
}

\providecommand{\abcellM}[2]{%
  \makebox[3.8em][r]{#1}%
  \makebox[2.2em][l]{\,{\tiny(#2\%)}}%
}
\providecommand{\abcellboldM}[2]{%
  \makebox[3.8em][r]{\textbf{#1}}%
  \makebox[2.2em][l]{\,{\tiny(#2\%)}}%
}
\providecommand{\abbaseM}[1]{%
  \makebox[3.8em][r]{#1}%
  \makebox[2.2em][l]{\,{}}%
}

\providecommand{\abcellT}[2]{%
  \makebox[3.5em][r]{#1}%
  \makebox[2.5em][l]{\,{\tiny(#2\%)}}%
}
\providecommand{\abcellboldT}[2]{%
  \makebox[3.5em][r]{\textbf{#1}}%
  \makebox[2.5em][l]{\,{\tiny(#2\%)}}%
}
\providecommand{\abbaseT}[1]{%
  \makebox[3.5em][r]{#1}%
  \makebox[2.5em][l]{\,{}}%
}

\centering
\caption{\textbf{Carryover-interface ablation.}
Qwen3-4B Markov diagnostic probe at $T=0$.
Cells report $\tau$ (gain over baseline, \%).}
\label{tab:ablations}

\footnotesize
\setlength{\tabcolsep}{5pt}
\renewcommand{\arraystretch}{0.95}

\begin{tabular}{l c c c}
\toprule
Variant & HumanEval & Math & MT-Bench \\
\midrule

Baseline (nothing carried)
& \abbaseH{3.10}
& \abbaseM{3.72}
& \abbaseT{2.46} \\

\addlinespace[2pt]
\multicolumn{4}{l}{\textit{(a) Carried-over information}} \\

\hspace{0.5em}Draft-emitted tokens
& \abcellH{3.05}{-1.8}
& \abcellM{3.76}{+1.2}
& \abcellT{2.41}{-2.4} \\

\hspace{0.5em}Target-emitted tokens
& \abcellH{3.16}{+1.8}
& \abcellM{3.93}{+5.8}
& \abcellT{2.52}{+2.3} \\

\hspace{0.5em}Target top-5 weighted
& \abcellH{3.18}{+2.7}
& \abcellM{3.92}{+5.4}
& \abcellT{2.56}{+4.0} \\

\rowcolor{carryshade}
\hspace{0.5em}Target hidden states (ours)
& \abcellboldH{3.41}{+10.0}
& \abcellboldM{4.27}{+14.8}
& \abcellboldT{2.70}{+9.6} \\

\midrule

\multicolumn{4}{l}{\textit{(b) Distinguishing carried-over hidden states}} \\

\hspace{0.5em}No explicit discrimination
& \abcellH{3.41}{+10.1}
& \abcellM{4.18}{+12.5}
& \abcellT{2.67}{+8.5} \\

\hspace{0.5em}Separate projector
& \abcellH{3.45}{+11.4}
& \abcellM{4.28}{+15.2}
& \abcellT{2.73}{+10.8} \\

\rowcolor{carryshade}
\hspace{0.5em}Carryover embedding (ours)
& \abcellboldH{3.41}{+10.0}
& \abcellboldM{4.27}{+14.8}
& \abcellboldT{2.70}{+9.6} \\

\bottomrule
\end{tabular}
\end{table}

\paragraph{What should be recycled?}
A rejected branch offers several forms of information for the next draft. Table~\ref{tab:ablations}(a) compares carrying over draft-emitted tokens, target-token predictions, and target hidden states. Token hints initialize the next proposal in place of \texttt{[MASK]} tokens. This Qwen3-4B Markov probe uses 6 training epochs on approximately 40K examples and evaluates HumanEval, Math, and MT-Bench. Draft-emitted token hints give inconsistent gains, while target-token hints yield modest improvements. Recycling target hidden states gives the largest gains among these designs. However, token hints enter through the initial proposal embeddings, whereas carried hidden states enter as KV context accessed by existing attention layers. Thus, this comparison changes both the information carried over and how it enters the drafter; it does not completely isolate their individual contributions.

\paragraph{How should rejected states be distinguished?}
Rejected hidden states describe an alternative continuation, motivating an explicit distinction from accepted context. Table~\ref{tab:ablations}(b) compares no explicit distinction, a separate projector, and the carryover embedding. Reusing rejected states without an explicit distinction from accepted context remains competitive, while the 19.7M-parameter projector gives slightly higher acceptance than the embedding. We choose the single embedding because it retains nearly the same acceptance without an additional projection module.

\section{How do rejected states contribute to drafting?}
\label{sec:analysis}
In this section, we first test the effect of rejected states at matched decoding positions and then examine why gains vary across tasks.

\subsection{Target-token recall at matched anchors}
\paragraph{Controlling the decoding point.} Even from the same context, different drafter rollouts can reach different verification boundaries. Therefore, we generate with Carryover ON and replay its draft starting points with 3 variants: Carryover ON, the same weights without carried-over states (OFF), and Baseline. All variants share committed tokens, accepted target hidden states, and target KV cache; only ON receives the recorded rejected states. Each proposal is scored under its own within-block prefix. We use the final Qwen3-4B Markov checkpoints on all 844 prompts at $T=0$.

\begin{figure*}[t]
\centering
\includegraphics[width=0.90\textwidth]{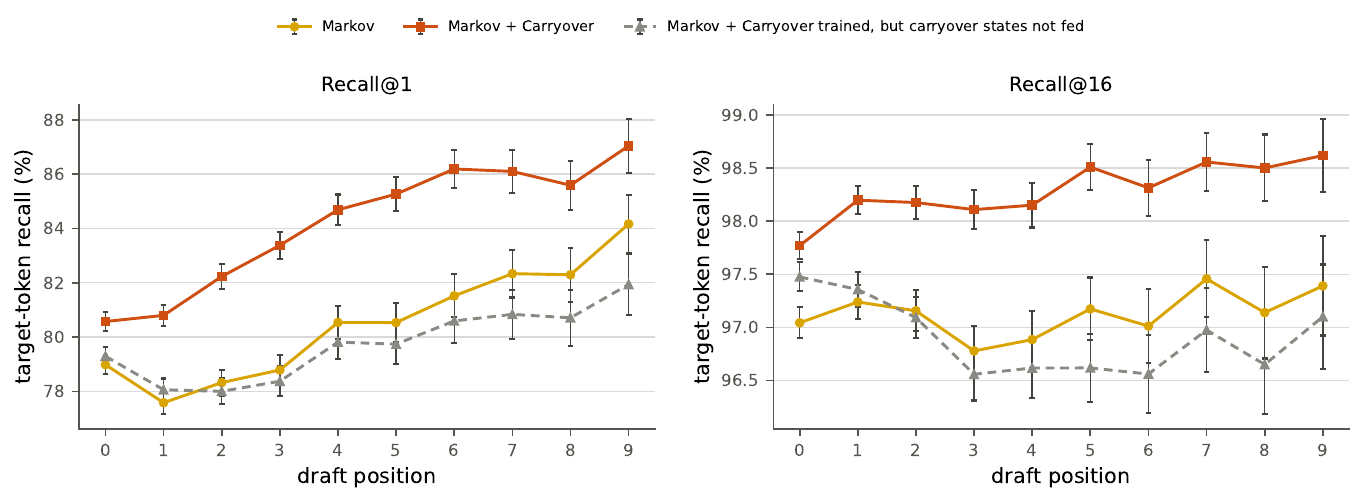}
\caption{\textbf{Target-token recall from identical decoding anchors.} Qwen3-4B Markov on 844 prompts at $T=0$: Baseline, Carryover with states (ON), and the same Carryover weights without them (OFF). Appendix~\ref{app:recall-protocol} details the sampling and error bars.}
\label{fig:recall}
\end{figure*}

Recall@$k$ measures how often the top target token appears among the top $k$ draft candidates. We report Recall@1 and Recall@16 to assess the top prediction and coverage of the candidate set, following the analysis in DFlash 2 \citep{incoai2026dflash2}. All variants use matched rounds at each position (Appendix~\ref{app:recall-protocol}).

\paragraph{Improved predictions from carried-over states.}
ON exceeds OFF at every displayed position by 1.28--5.60 Recall@1 points and 0.29--1.89 Recall@16 points (Figure~\ref{fig:recall}). Supplying rejected states thus improves both the top prediction and coverage of the target token within the candidate set. Gains begin at position 0, where all variants share the same accepted anchor, and persist across the block.

ON--OFF indirectly tests the learned use of rejected states: OFF replays ON starting points rather than its own rollout and withholds context the model was trained to use. Appendix~\ref{app:attention} complements this diagnostic by showing attention to carried states (Figure~\ref{fig:attention}).

\subsection{Why do gains differ across datasets?}
\label{sec:reconvergence}
Carryover improves acceptance length by different amounts across datasets, reaching 28.6\% on translation at $T=1$. To investigate this variation, we examine which properties of rejected states are associated with larger gains and how these properties differ across datasets. At $T=0$, we compare rejected-state counts, consecutive and total proposal--verifier matches after rejection, and \emph{re-convergence}. State count measures how much context is available; acceptance after rejection measures how often draft tokens are accepted along the rejected trajectory. Re-convergence measures whether predictions from the rejected trajectory match the eventual continuation. For clarity, we define the re-convergence count as follows. For a rejected suffix of length $L$, let $z_q$ be the verifier's top-1 next-token prediction from its $q$th rejected state, and $y_q$ the eventually committed token at the corresponding predicted position. We count
\begin{equation}
R=\sum_{q=0}^{L-1}\mathbf{1}\!\left[\exists\,\delta\in\{-1,0,+1\}:z_{q+\delta}=y_q\right],
\label{eq:reconvergence}
\end{equation}
ignoring out-of-range indices. The one-position tolerance allows a local insertion or deletion. $R$ measures retrospective agreement with the eventual continuation; it is never used to select carried-over states or supervise training.

\begin{figure*}[t]
\centering
\includegraphics[width=0.97\textwidth]{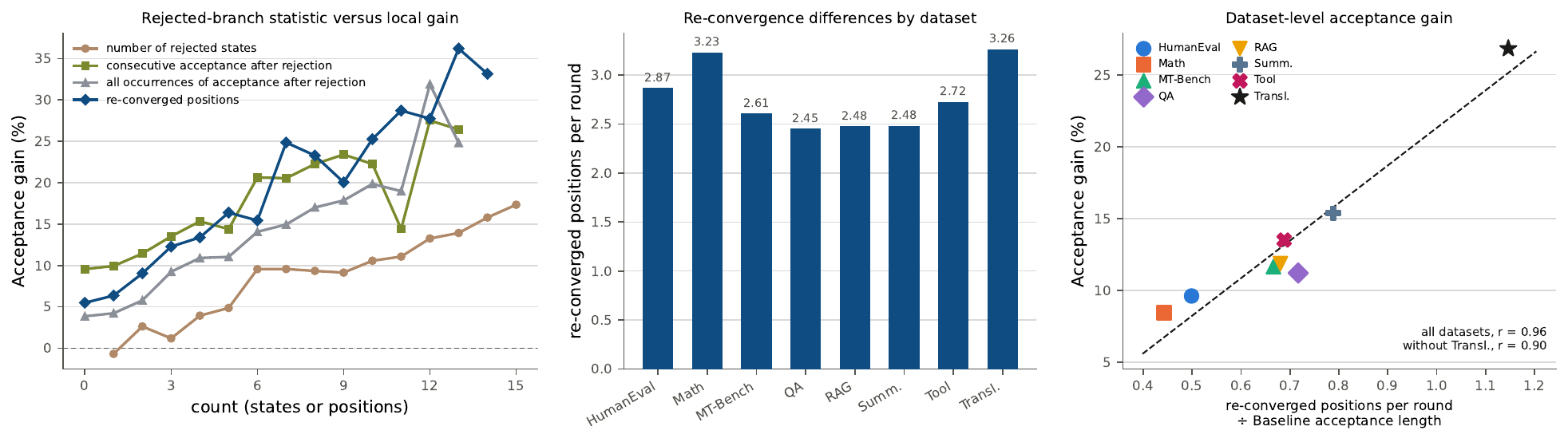}
\caption{\textbf{Rejected-state statistics and Carryover gains.} Final Qwen3-4B Markov at $T=0$. Left: local replay gain grouped by rejected-branch statistics. Center: re-converged positions per rejected round. Right: normalized re-convergence versus final dataset-level acceptance gain.}
\label{fig:reconvergence}
\end{figure*}

Figure~\ref{fig:reconvergence} shows that Carryover gains are more closely
associated with re-convergence than with rejected-state count or
post-rejection acceptance (Appendix~\ref{app:reconvergence}). We distinguish
post-rejection acceptance from re-convergence: a token may be accepted along
the rejected branch yet differ from the token eventually committed by the
target. Altogether, these results suggest that the usefulness of rejected
states depends more on their predictiveness of the target's eventual
continuation than on either post-rejection acceptance or rejected-suffix
length.

Across drafters, we also find that Markov produces more re-converged positions than DFlash on all 8 datasets (2.76 versus 2.51 per rejected round) and obtains larger Carryover gains on 7 of 8 (Appendix~\ref{app:family-reconvergence}). Higher re-convergence thus offers one explanation for the larger gains of Markov. It also suggests that stronger drafters leave more useful rejected states, so the benefit of Carryover can grow as drafters improve.

\section{Discussion}
\label{sec:future}
\paragraph{Confidence scheduling with Carryover.}
DSpark uses estimated acceptance probabilities together with hardware capacity to adapt the verification length \citep{cheng2026dspark}. Carryover is compatible in principle with adaptive verification, but the scheduling policy changes which rejected states are available to the next draft. A joint evaluation would therefore require training Carryover under the same scheduling policy used at inference, so that its rejected-state inputs match the deployment distribution. In addition, because Carryover changes the drafter's predictions, the confidence estimates and scheduling thresholds may require recalibration. We therefore do not evaluate Carryover with confidence scheduling in this work.

\paragraph{Tree decoding with Carryover.}
Tree verification methods such as Medusa and DDTree spend more compute on alternative paths to increase acceptance length \citep{cai2024medusa,ringel2026ddtree}. We anticipate the combination of Carryover with tree verification to be especially interesting: the additional verification produces both more rejected states and different information across branches, which gives subsequent drafts more opportunity for reuse. It also introduces an
additional design question: which branch states should be retained for the
next draft? Applying Carryover to tree decoding would therefore require
matching the tree-construction policy between training and inference and
determining how states from multiple rejected branches should be selected or
combined. We leave this extension to future work.

\paragraph{On-policy training with rollouts.}
Our parallel draft--verify--draft procedure is designed to obtain
inference-aligned rejected states while retaining parallel training on recorded
responses. When training already includes target-verified rollouts, however,
rejected states arise naturally during verification. For example, Draft-OPD
uses target-assisted rollouts and replays drafting at verification-exposed
error positions \citep{lei2026draftopd}. Carryover could reuse the hidden states
produced by these verification steps without requiring an additional target
verification stage. Such rollouts would also make it possible to train Carryover with retention beyond a single round, which our parallel training does not cover.

\section{Related work}
\label{sec:related}
\paragraph{Target-conditioned drafting.}
EAGLE and EAGLE-3 use target hidden states for autoregressive drafting, while DFlash supplies them as KV context for parallel block prediction \citep{li2024eagle,li2025eagle3,chen2026dflash}. Domino, DSpark, xPress, and DFlash 2 strengthen dependencies or refine candidates within a block; D-PACE changes the training objective \citep{huang2026domino,cheng2026dspark,wang2026xpress,incoai2026dflash2,wu2026dpace}. Carryover extends the target hidden states available after verification, complementing improvements to how the current block is generated.

\paragraph{Recycling speculative computation.}
Token Recycling caches target candidates, Lookahead reuses Jacobi n-grams, and Ouroboros reuses phrases \citep{luo2025tokenrecycling,fu2024lookahead,zhao2024ouroboros}. At the representation level, SD$^2$ learns a verifier-derived steering vector for an independently pretrained autoregressive drafter and injects it through MLPs \citep{berdoz2026sd2}. Lyanna recycles drafter hidden states into token trees \citep{chen2026makeeverydraftcount}. Carryover instead retains an ordered sequence of rejected target hidden states and supplies them through the existing KV interface of a target-conditioned drafter.

\section{Conclusion}
Carryover Drafting recycles rejected target hidden states through the drafter's existing KV interface with one learned embedding and one round of additional context. Parallel draft--verify--draft training exposes these states without sequential rollouts. Across different target models and drafting regimes, Carryover consistently improves acceptance length and end-to-end speedup.

\clearpage
\typeout{CARRYOVER-MAIN-LAST-PAGE=\number\numexpr\value{page}-1\relax}
\bibliography{paper/iclr2027/references_carryover}

@inproceedings{leviathan2023fast,
  title={Fast Inference from Transformers via Speculative Decoding},
  author={Leviathan, Yaniv and Kalman, Matan and Matias, Yossi},
  booktitle={International Conference on Machine Learning},
  year={2023}
}

@article{chen2023accelerating,
  title={Accelerating Large Language Model Decoding with Speculative Sampling},
  author={Chen, Charlie and Borgeaud, Sebastian and Irving, Geoffrey and
          Lespiau, Jean-Baptiste and Sifre, Laurent and Jumper, John},
  journal={arXiv preprint arXiv:2302.01318},
  year={2023}
}

@inproceedings{cai2024medusa,
  title={Medusa: Simple {LLM} Inference Acceleration Framework with Multiple
         Decoding Heads},
  author={Cai, Tianle and Li, Yuhong and Geng, Zhengyang and Peng, Hongwu and
          Lee, Jason D. and Chen, Deming and Dao, Tri},
  booktitle={International Conference on Machine Learning},
  pages={5209--5235},
  year={2024}
}

@inproceedings{li2024eagle,
  title={{EAGLE}: Speculative Sampling Requires Rethinking Feature
         Uncertainty},
  author={Li, Yuhui and Wei, Fangyun and Zhang, Chao and Zhang, Hongyang},
  booktitle={International Conference on Machine Learning},
  year={2024}
}

@inproceedings{li2025eagle3,
  title={{EAGLE-3}: Scaling up Inference Acceleration of Large Language
         Models via Training-Time Test},
  author={Li, Yuhui and Wei, Fangyun and Zhang, Chao and Zhang, Hongyang},
  booktitle={Advances in Neural Information Processing Systems},
  year={2025}
}

@inproceedings{chen2026dflash,
  title={{DFlash}: Block Diffusion for Flash Speculative Decoding},
  author={Chen, Jian and Liang, Yesheng and Liu, Zhijian},
  booktitle={International Conference on Machine Learning},
  year={2026}
}

@article{cheng2026dspark,
  title={{DSpark}: Confidence-Scheduled Speculative Decoding with
         Semi-Autoregressive Generation},
  author={Cheng, Xin and Yu, Xingkai and Shao, Chenze and Li, Jiashi and
          Xiong, Yunfan and Qian, Yi and Zhu, Jiaqi and Ma, Shirong and
          Zhang, Xiaokang and Ye, Jiasheng and others},
  journal={arXiv preprint arXiv:2607.05147},
  year={2026}
}

@article{huang2026domino,
  title={{Domino}: Decoupling Causal Modeling from Autoregressive Drafting in
         Speculative Decoding},
  author={Huang, Jianuo and Zhang, Yaojie and Zhang, Qituan and Lin, Hao and
          Xu, Hanlin and Zhang, Linfeng},
  journal={arXiv preprint arXiv:2605.29707},
  year={2026}
}

@article{wang2026xpress,
  title={{xPress}: Parallel Refinement for Diffusion Drafters in Speculative
         Decoding},
  author={Wang, Zheng and Wertheimer, Davis and Lim, Yu Chin Fabian and
          Srivatsa, Mudhakar and Ganti, Raghu K. and Zhang, Minjia and Wang,
          Naigang},
  journal={arXiv preprint arXiv:2608.02438},
  year={2026}
}

@article{wu2026dpace,
  title={{D-PACE}: Dynamic Position-Aware Cross-Entropy for Parallel
         Speculative Drafting},
  author={Wu, Tianyu and Yao, Yu and Qi, Zhenting and Zheng, Han and Wang,
          Zhuohan and Ma, Haoran and Liao, Lawrence and Lakkaraju,
          Himabindu and Li, Ju and Du, Yilun},
  journal={arXiv preprint arXiv:2605.18810},
  year={2026}
}

@inproceedings{zhao2024ouroboros,
  title={Ouroboros: Generating Longer Drafts Phrase by Phrase for Faster
         Speculative Decoding},
  author={Zhao, Weilin and Huang, Yuxiang and Han, Xu and Xu, Wang and Xiao,
          Chaojun and Zhang, Xinrong and Fang, Yewei and Zhang, Kaihuo and
          Liu, Zhiyuan and Sun, Maosong},
  booktitle={Empirical Methods in Natural Language Processing},
  pages={13378--13393},
  year={2024},
  doi={10.18653/v1/2024.emnlp-main.742}
}

@inproceedings{luo2025tokenrecycling,
  title={Turning Trash into Treasure: Accelerating Inference of Large
         Language Models with Token Recycling},
  author={Luo, Xianzhen and Wang, Yixuan and Zhu, Qingfu and Zhang, Zhiming
          and Zhang, Xuanyu and Yang, Qing and Xu, Dongliang},
  booktitle={Annual Meeting of the Association for Computational Linguistics},
  pages={6816--6831},
  year={2025},
  doi={10.18653/v1/2025.acl-long.338}
}

@inproceedings{fu2024lookahead,
  title={Break the Sequential Dependency of {LLM} Inference Using Lookahead
         Decoding},
  author={Fu, Yichao and Bailis, Peter and Stoica, Ion and Zhang, Hao},
  booktitle={International Conference on Machine Learning},
  year={2024}
}

@inproceedings{berdoz2026sd2,
  title={Steering Pretrained Drafters During Speculative Decoding},
  author={Berdoz, Fr{\'e}d{\'e}ric and Rheinboldt, Peer and Wattenhofer,
          Roger},
  booktitle={AAAI Conference on Artificial Intelligence},
  pages={30067--30075},
  year={2026},
  doi={10.1609/aaai.v40i36.40255}
}

@misc{redhat2026speculators,
  title={Speculator Benchmarks},
  author={{Red Hat AI}},
  year={2026},
  howpublished={Hugging Face dataset},
  url={https://huggingface.co/datasets/RedHatAI/speculator_benchmarks},
  note={Revision 2ae86affa2cb97a972b7fc681dd51c04fbff083e}
}

@article{yang2025qwen3,
  title={Qwen3 Technical Report},
  author={Yang, An and Li, Anfeng and Yang, Baosong and Zhang, Beichen and
          Hui, Binyuan and Zheng, Bo and Yu, Bowen and Gao, Chang and Huang,
          Chengen and Lv, Chenxu and others},
  journal={arXiv preprint arXiv:2505.09388},
  year={2025}
}

@inproceedings{kwon2023vllm,
  title={Efficient Memory Management for Large Language Model Serving with
         {PagedAttention}},
  author={Kwon, Woosuk and Li, Zhuohan and Zhuang, Siyuan and Sheng, Ying and
          Zheng, Lianmin and Yu, Cody Hao and Gonzalez, Joseph E. and Zhang,
          Hao and Stoica, Ion},
  booktitle={ACM SIGOPS Symposium on Operating Systems Principles},
  year={2023}
}

@article{lei2026draftopd,
  title={{Draft-OPD}: On-Policy Distillation for Speculative Draft Models},
  author={Lei, Haodi and Li, Yafu and Zhang, Haoran and Zhang, Shunkai and
          Cheng, Qianjia and Qu, Xiaoye and Cui, Ganqu and Zhou, Bowen and
          Ding, Ning and Luo, Yun and Cheng, Yu},
  journal={arXiv preprint arXiv:2605.29343},
  year={2026}
}

@article{gemmateam2026gemma4,
  title={Gemma 4 Technical Report},
  author={{Gemma Team}},
  journal={arXiv preprint arXiv:2607.02770},
  year={2026}
}

@article{chen2026makeeverydraftcount,
  title={Make Every Draft Count: Hidden State based Speculative Decoding},
  author={Chen, Yuetao and Wang, Xuliang and Zheng, Xinzhou and Li, Ming and
          Wang, Peng and Xu, Hong},
  journal={arXiv preprint arXiv:2602.21224},
  year={2026}
}

@article{ringel2026ddtree,
  title={Accelerating Speculative Decoding with Block Diffusion Draft Trees},
  author={Ringel, Liran and Romano, Yaniv},
  journal={arXiv preprint arXiv:2604.12989},
  year={2026}
}

@misc{incoai2026dflash2,
  author       = {{Inco AI}},
  title        = {DFlash 2: Keep Drafting Parallel},
  year         = {2026},
  month        = aug,
  howpublished = {\url{https://inco.ai/blog/dflash2/}}
}

@inproceedings{ding2023ultrachat,
  title={Enhancing Chat Language Models by Scaling High-Quality
         Instructional Conversations},
  author={Ding, Ning and Chen, Yulin and Xu, Bokai and Qin, Yujia and Zheng,
          Zhi and Hu, Shengding and Liu, Zhiyuan and Sun, Maosong and Zhou,
          Bowen},
  booktitle={Conference on Empirical Methods in Natural Language Processing},
  pages={3029--3051},
  year={2023},
  doi={10.18653/v1/2023.emnlp-main.183}
}

@inproceedings{xu2024magpie,
  title={Magpie: Alignment Data Synthesis from Scratch by Prompting Aligned
         {LLM}s with Nothing},
  author={Xu, Zhangchen and Jiang, Fengqing and Niu, Luyao and Deng, Yuntian
          and Poovendran, Radha and Choi, Yejin and Lin, Bill Yuchen},
  booktitle={International Conference on Learning Representations},
  year={2025},
  url={https://arxiv.org/abs/2406.08464}
}

@misc{amazon2026peaglecard,
  title={{GPT-OSS-120B P-EAGLE} Model Card},
  author={{Amazon Web Services}},
  year={2026},
  howpublished={Hugging Face model card},
  url={https://huggingface.co/amazon/gpt-oss-120b-p-eagle},
  note={Documents the UltraChat-200K and
        Magpie-Llama-3.1-Pro-300K-Filtered prompt mixture}
}
\bibliographystyle{iclr2027_conference}
\clearpage
\appendix
\AtBeginEnvironment{enumerate}{\setlength{\itemsep}{2pt}\setlength{\parsep}{0pt}}
\section{Inference and training algorithms}
\label{app:algorithms}
Algorithms~\ref{alg:inference} and~\ref{alg:training} compare the baseline and Carryover directly. At inference, the change is to collect all proposal states, add the learned embedding only to the rejected portion, and apply the existing projections. At training time, the same fixed response supplies supervision at anchors shifted by a parallel draft--verify step. The added embedding is shown in orange.

\subsection{Inference update}
\begin{algorithm}[H]
\caption{Baseline and Carryover: the post-verification update}
\label{alg:inference}
\small
\textbf{Given:} target features $H_{1:B}$ at the proposed positions, accepted length $m$, and existing committed drafter KV $C_\ell$. Both methods use the usual draft--verify procedure and commit the accepted prefix plus correction or bonus with the usual target KV update.
\par\medskip
\begin{minipage}[t]{0.46\linewidth}
\textbf{Baseline: collect accepted states}
\begin{enumerate}
\item $H^{\mathrm{keep}}\gets H_{1:m}$
\item $Z\gets F(H^{\mathrm{keep}})$
\item $J_\ell\gets(K_\ell(Z),V_\ell(Z))$
\item $C_\ell\gets[C_\ell;J_\ell]$
\item $\mathrm{ctx}_\ell\gets C_\ell$
\end{enumerate}
\end{minipage}\hfill
\begin{minipage}[t]{0.51\linewidth}
\textbf{Carryover: collect all states}
\begin{enumerate}
\item $H^{\mathrm{keep}}\gets H_{1:B}$
\item $H^{\mathrm{keep}}_{m+1:B}\gets H^{\mathrm{keep}}_{m+1:B}+\textcolor{orange}{E_{\mathrm{carry}}}$
\item $Z\gets F(H^{\mathrm{keep}})$
\item $J_\ell\gets(K_\ell(Z),V_\ell(Z))$
\item $C_\ell\gets[C_\ell;J_{\ell,1:m}]$
\item $U_\ell\gets J_{\ell,m+1:B}$ \hfill (Replace old $U_\ell$)
\item $\mathrm{ctx}_\ell\gets[C_\ell;U_\ell]$
\end{enumerate}
\end{minipage}
\par\medskip
Apply the KV steps at every drafter layer $\ell$. The next draft reads $\mathrm{ctx}_\ell$. Initialize $U_\ell=\varnothing$; full acceptance also leaves it empty.
\end{algorithm}

\paragraph{Cache semantics.} The two columns show the same feature-processing path, with a different choice of retained rows. $F$ is the existing projection and $K_\ell,V_\ell$ are the layer-specific key and value projections; $J_\ell$ denotes their output KV rows. Accepted rows extend committed drafter context $C_\ell$. Rejected rows are exposed only as read-only drafter context $U_\ell$ for one next proposal. They are collected before the verification workspace is reused and never enter the target's committed KV cache. Stopping rules and output truncation follow the baseline. Each rejected state keeps the absolute position of the token at which it was computed, and RoPE is applied once, when the drafter projects it into keys.
\paragraph{Why verification remains valid.} Conditioned on the decoding history and carried input, the drafter defines a proposal distribution $q$. Applying the usual target acceptance and residual correction with that actual $q$ preserves the target distribution, just as for other history-dependent proposals. Carryover changes this proposal distribution but neither the target probabilities nor the acceptance rule. Rejected text remains an input to the drafter only; allowing it to overwrite committed target KV would violate that separation.

\clearpage
\subsection{Parallel training update}
\begin{algorithm}[H]
\caption{Parallel training: shared setup and alternative update rules}
\label{alg:training}
\small
\textbf{Given:} one fixed target-sampled response $y$ per prompt and its recorded target features; frozen target $P$; drafter $D_\theta$; proposal length $B$; baseline loss $\mathcal{L}_{\mathrm{base}}$. The prompt is implicit in each prefix of $y$.

\textbf{Existing interface:} $\Phi_\ell(H)=(K_\ell(F(H)),V_\ell(F(H)))$ at every drafter layer $\ell$. Only the additional embedding $E_{\mathrm{carry}}$ is shown in orange.

\textbf{Shared setup}
\begin{enumerate}
\item $\{a_i\}\gets\operatorname{RandomAnchors}(y,\mathrm{count}=512)$ \hfill (Shared random-anchor sampler)
\item Use recorded prefixes, target features, and continuation labels from the same $y$.
\end{enumerate}
Apply either the Baseline update or the Carryover update below.

\textbf{Baseline update} \quad With gradients; all anchors in parallel
\begin{enumerate}
\item $\widehat y_i\gets\operatorname{Draft}(D_\theta,y_{\leq a_i})$
\item Update $\theta$ with $\mathcal{L}_{\mathrm{base}}$ against recorded continuations $y_{>a_i}$.
\end{enumerate}
\textbf{Carryover update}

\emph{Draft--verify: no gradients; all initial anchors in parallel}
\begin{enumerate}
\item $d_i\gets\operatorname{GreedyDraft}(D_\theta,y_{\leq a_i})$
\item $(m_i,H_i)\gets\operatorname{GreedyVerify}(P,y_{\leq a_i},d_i)$ \hfill (Frozen target; isolated branches)
\item $a'_i\gets a_i+m_i+1;\quad H_i^{\mathrm{rej}}\gets\operatorname{sg}(H_{i,m_i+1:B})$
\item Keep every sampled anchor; if $m_i=B$, set $H_i^{\mathrm{rej}}=\varnothing$ and use empty carried-over context.
\end{enumerate}
\emph{Second draft: with gradients; all sampled anchors in parallel}
\begin{enumerate}\setcounter{enumi}{4}
\item $U_{i,\ell}\gets\Phi_\ell(H_i^{\mathrm{rej}}+\textcolor{orange}{E_{\mathrm{carry}}})$
\item $\widehat y_i\gets\operatorname{Draft}(D_\theta,y_{\leq a'_i},U_i)$ \hfill (Recorded prefix and features)
\item Update $\theta$ and $E_{\mathrm{carry}}$ with $\mathcal{L}_{\mathrm{base}}$ against recorded continuations $y_{>a'_i}$.
\end{enumerate}
\end{algorithm}

\paragraph{One fixed response, shifted anchors.}
Both procedures reuse the same target-sampled response $y$ and randomly sampled anchors, following parallel training in DFlash and DSpark \citep{chen2026dflash,cheng2026dspark}. Our experiments sample 512 initial anchors per response. Here $a_i$ denotes an anchor and $m_i$ the number of accepted proposals. Carryover moves each anchor to $a'_i=a_i+m_i+1$, the next correction or bonus position on that same response. The second draft reads the recorded prefix and recorded target features at $a'_i$ and learns its recorded continuation. Verification supplies rejected-state inputs and the shift; it does not sample a new target response or replace the labels with a newly generated continuation. Each initial anchor supplies 1 supervised draft, including when the proposal is fully accepted.

\paragraph{Parallelism and gradients.} The draft--verify--draft stages are sequential within each local transition, but all anchors are processed in parallel within a stage. A block-diagonal causal mask isolates target branches: each reads its recorded prefix and its own proposed predecessors. The target partial prefill may be chunked without making one anchor depend on another. The first drafter pass and target verification are detached. The second drafter pass, conditioned on carried-over states when present, trains the drafter, shared projections $\Phi_\ell$ from Equation~\ref{eq:features}, and embedding using the corresponding baseline objective.
\paragraph{Fully accepted proposals.} All sampled anchors contribute to the second-stage loss. When the first proposal is fully accepted, $H_i^{\mathrm{rej}}=\varnothing$ and the second drafter pass uses only the recorded context at the shifted anchor. This is the corresponding baseline training procedure at the shifted anchor, so fully accepted proposals still provide supervision.
\clearpage
\section{Training and evaluation protocol}
\subsection{Training data and settings}\label{app:training-settings}
\begin{table}[H]
\centering
\caption{\textbf{Training settings.} Shared architecture and optimizer settings in the final release configuration; Carryover continues the corresponding Baseline-10 checkpoint. Target-layer IDs are zero-based transformer-block indices.}
\label{tab:training}
\small
\begin{tabular}{lcc}
\toprule
Setting & Qwen3-4B & Gemma 4 12B \\
\midrule
Target feature layers & 1, 9, 17, 25, 33 & 5, 17, 29, 41, 46 \\
Drafter layers / proposed tokens & 5 / 15 & 5 / 15 \\
Markov correction rank & 256 & 256 \\
Initial anchors / response & 512 & 512 \\
Maximum training sequence length & 4,096 & 4,096 \\
Global batch size & 512 & 512 \\
Learning rate / warmup ratio & $6\times10^{-4}$ / 0.04 & $6\times10^{-4}$ / 0.04 \\
Precision / gradient clipping & bfloat16 / 1.0 & bfloat16 / 1.0 \\
Baseline / Carryover training epochs & 10 / 3 & 10 / 3 \\
\bottomrule
\end{tabular}
\end{table}

\paragraph{Recorded supervision.} The 500K-prompt mixture comprises UltraChat-200K and Magpie-Llama-3.1-Pro-300K-Filtered \citep{ding2023ultrachat,xu2024magpie}. Each target regenerates one response per prompt following the public speculator recipe \citep{amazon2026peaglecard}; its Baseline and Carryover variants share these responses.
\paragraph{Shared settings.} Baseline and Carryover use the same architecture, target-feature layers, initial-anchor sampler (512 per response), and loss. The second drafter pass retains the objective of the corresponding baseline. As in DFlash and DSpark, loss weights decay exponentially with position inside the draft block, prioritizing early predictions because rejection stops acceptance of the suffix \citep{chen2026dflash,cheng2026dspark}. DFlash uses position-weighted cross-entropy; the Markov baseline additionally uses distribution-matching and confidence objectives from the DSpark training recipe. Disabling confidence scheduling affects the inference policy, not these training objectives.
\paragraph{Continuation.} Carryover continues the 10-epoch baseline checkpoint for 3 epochs (Table~\ref{tab:training}), with fresh optimizer and scheduler states. Its second-stage loss uses the shifted anchors in Algorithm~\ref{alg:training}. Detached proposal generation and frozen-target partial prefill add computation while preserving parallelism across anchors. The recorded target responses remain fixed.

\clearpage
\subsection{Evaluation protocols}\label{app:evaluation}
\begin{table}[H]
\centering
\caption{\textbf{Evaluation protocols.} Offline and online measurements answer different questions and are not averaged together. All output caps count generated tokens.}
\label{tab:evaluation}
\small
\setlength{\tabcolsep}{3pt}
\begin{tabularx}{\textwidth}{@{}lXXX@{}}
\toprule
 & Dataset results & Average speedup & Concurrency sweep \\
\midrule
Engine & Offline vLLM & Online vLLM & Online vLLM \\
Workload & Each dataset & Mixed 844 prompts & Mixed 844 prompts \\
Batch / client concurrency & Batch size 1 & Concurrency 1 & 1, 2, 4, 8, 16, 32 \\
Output cap & 2,048 & 1,024 & 1,024 \\
Temperature & 0 and 1 & 0 and 1 & 0 \\
Reported quantity & $\tau$, speedup vs. AR & Speedup vs. AR & Output tokens/s \\
Hardware & 1 H200 & 1 H200 & 1 H200 \\
\bottomrule
\end{tabularx}
\end{table}

\paragraph{Benchmark and decoding.} The Red Hat speculator benchmark comprises 844 prompts: 164 for HumanEval, 200 for tool calling, and 80 each for Math, MT-Bench, QA, RAG, summarization, and translation \citep{redhat2026speculators}. Dataset-level and suite-speedup measurements cover greedy decoding ($T=0$) and stochastic decoding ($T=1$); the concurrency sweep uses $T=0$.
\paragraph{Execution and output limits.} All measurements use vLLM on 1 H200 GPU \citep{kwon2023vllm}. Dataset rows use offline inference at batch size 1 with a 2,048-token output cap. Average speedup is measured separately using online serving at client concurrency 1 on the mixed 844-prompt workload with a 1,024-token cap. The throughput sweep uses the same online workload and cap at concurrency 1, 2, 4, 8, 16, 32. All caps count generated tokens. Evaluation settings are identical within each baseline--Carryover pair. The Avg. speedup entries in Table~\ref{tab:big} are these separate measurements on the combined workload, not arithmetic means of the offline dataset speedups.
\paragraph{Metric aggregation.} Acceptance length includes bonus tokens from verification and is micro-averaged over all verification rounds in a dataset. For online throughput, total generated output tokens are divided by the post-warmup serving makespan. Average speedup uses the separately timed autoregressive baseline on the same online workload. Displayed values are rounded after aggregation, so recomputing percentage gains from the printed values can differ slightly from the reported percentages.

\paragraph{Matched-anchor recall protocol.}\label{app:recall-protocol}
At draft position $j$, all 3 variants are evaluated on the same rounds, namely those in which ON and Baseline match the target through position $j-1$; OFF is included even if its earlier prediction differs. The set of rounds therefore changes with position. Recall is micro-averaged over these rounds, and error bars use a normal approximation.

\clearpage\section{Attention to carried states}\label{app:attention}
\begin{figure}[H]
\centering
\includegraphics[width=0.95\textwidth]{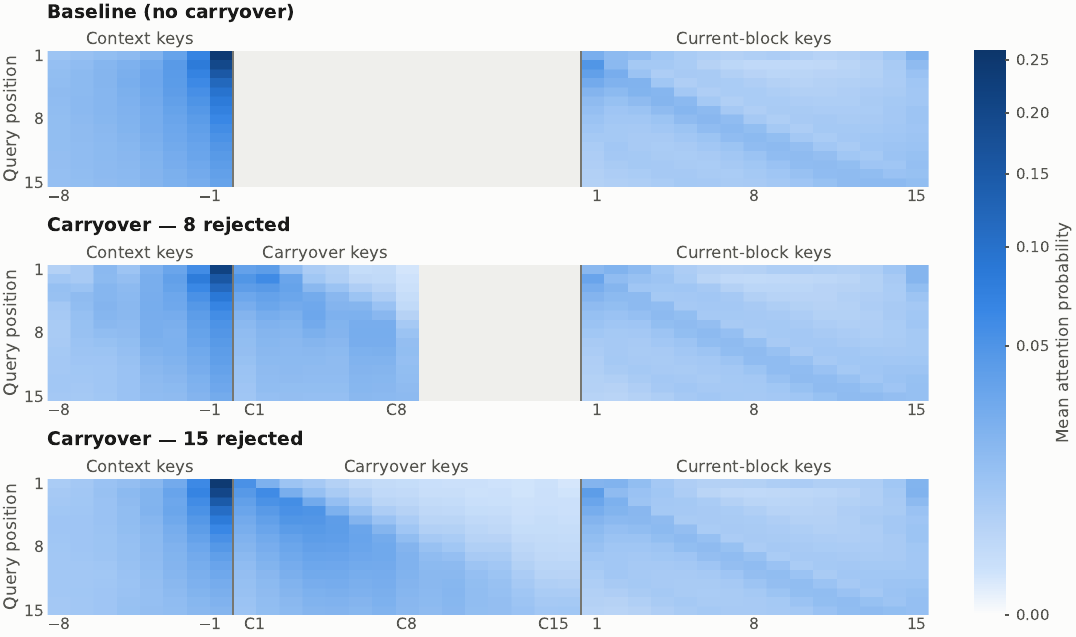}
\caption{\textbf{Attention diagnostic.} Visualized attention for DFlash checkpoints on MT-Bench: baseline, and Carryover rounds with 8 or 15 rejected states. Probabilities are averaged across heads, drafter layers, and rounds in each subset. The display crops context positions without renormalizing the probabilities.}
\label{fig:attention}
\end{figure}

Figure~\ref{fig:attention} shows how much attention the drafter places on carried rows. 19.78\% of full-softmax attention falls on carried rows. The 15-rejected subset assigns 23.40\% to those rows. The diagnostic establishes that the carried over interface is read and that attention varies with position. 

\Needspace{0.52\textheight}
\section{Re-convergence and acceptance after rejection}\label{app:reconvergence}

\begin{figure}[H]
\centering
\includegraphics[width=0.86\textwidth]{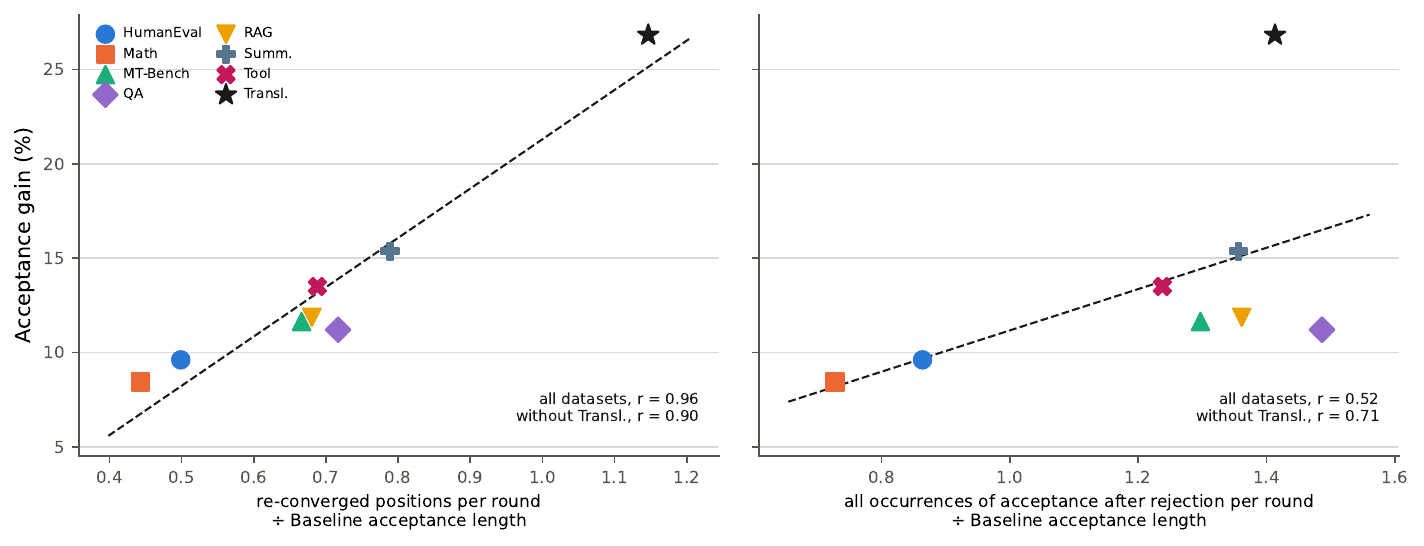}
\caption{\textbf{Agreement with the eventual continuation versus acceptance after rejection.} 8-dataset associations for final Qwen3-4B Markov. The left statistic counts re-converged positions; the right counts all local proposal--verifier matches after rejection. Both are normalized by the dataset's Baseline acceptance length.}
\label{fig:reconverged-vs-accepted}
\end{figure}

A draft token may be accepted later within its own rejected branch without matching the eventual continuation. Figure~\ref{fig:reconverged-vs-accepted} distinguishes these notions. Normalized re-convergence correlates with final dataset-level gain at $r=0.96$ (0.90 without Translation), compared with 0.52 (0.71) for all acceptance after rejection. For comparison, rejected-state count normalized by Baseline acceptance gives correlations of $r=0.74$ across all datasets and 0.84 without Translation.
\Needspace{0.43\textheight}
\section{Re-convergence across drafter types}\label{app:family-reconvergence}

\begin{figure}[H]
\centering
\includegraphics[width=1.0\textwidth]{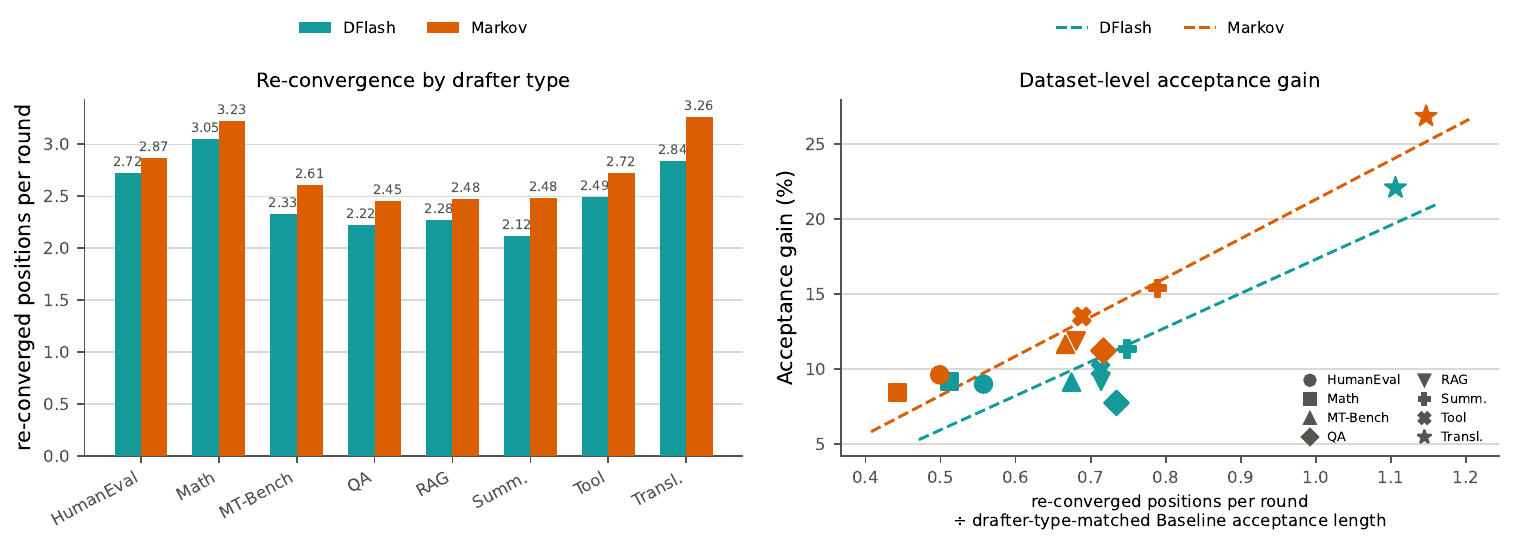}
\caption{\textbf{Rejected-branch structure across drafters.} Final Qwen3-4B DFlash and Markov pairs at $T=0$. Left: re-converged positions per rejected round. Right: the normalized statistic and final acceptance gain for each dataset and drafter.}
\label{fig:family-reconvergence}
\end{figure}

Markov averages 2.76 re-converged positions versus 2.51 for DFlash, a 10.2\% difference. It has higher re-convergence and baseline acceptance on every dataset, and higher Carryover gain on 7 of 8 datasets.

\Needspace{0.43\textheight}
\section{Acceptance across tasks and temperatures}\label{app:acceptance-t1}

\begin{figure}[H]
\centering
\includegraphics[width=1.0\textwidth]{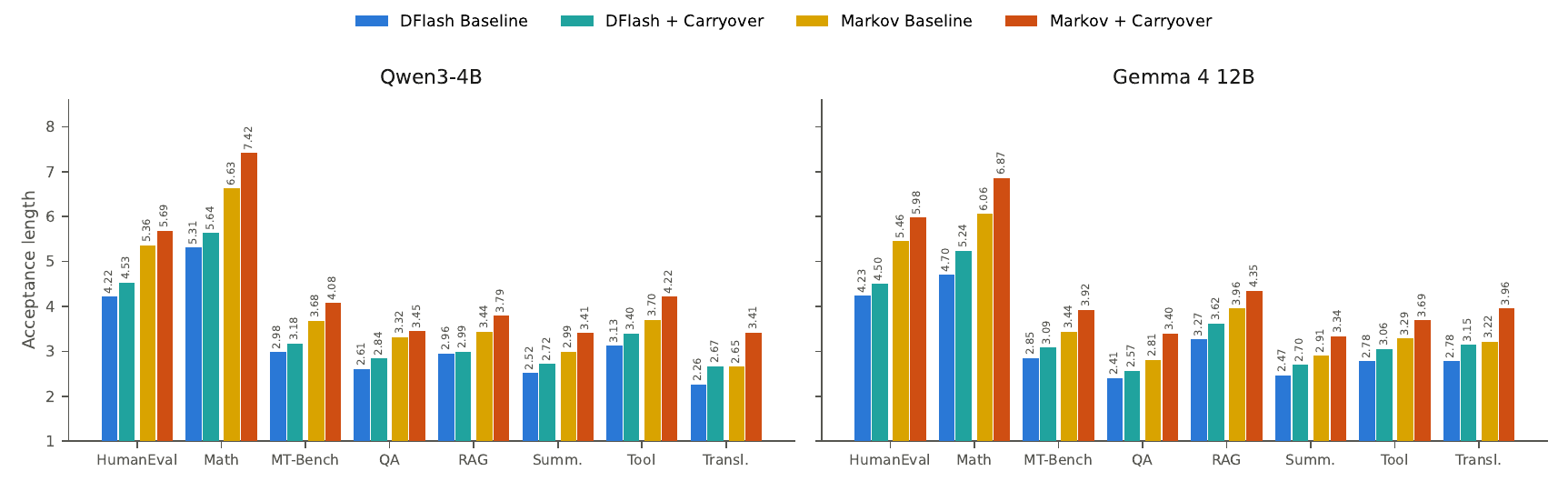}
\caption{\textbf{Acceptance length at $T=1$.} Baseline and Carryover for both targets and drafting regimes. These bars complement the greedy results in Figure~\ref{fig:acceptance}; the numerical results are in Table~\ref{tab:big}.}
\label{fig:acceptance-t1}
\end{figure}

The gain remains positive in every displayed stochastic setting, although its magnitude varies. For example, Qwen3-4B DFlash on RAG has a modest 1.1\% gain, whereas Qwen3-4B Markov on translation gains 28.6\%.

\subsection*{AI use statement}
Generative AI tools assisted source inspection, result collation, drafting, and formatting. The authors are responsible for reviewing the final claims and artifacts.

\end{document}